\documentclass{article}
\usepackage{ijcai26}

\usepackage{times}
\usepackage{soul}
\usepackage{url}
\usepackage[hidelinks]{hyperref}
\usepackage[utf8]{inputenc}
\usepackage[small]{caption}
\usepackage{graphicx}
\usepackage{amsmath}
\usepackage{amsthm}
\usepackage{booktabs}
\usepackage{algorithm}
\usepackage{algorithmic}
\usepackage[switch]{lineno}
\usepackage{cuted}
\usepackage{makecell}
\usepackage{colortbl}
\usepackage{pifont}
\usepackage{multirow}
\usepackage{amssymb}

\title{VGA-BenchV2: An Expanded Unified Benchmark and Multi-Model Framework for Evaluating Video Aesthetics and Generation Quality}

\author{
Longteng Jiang$^1$
\and
Dandan Zheng$^1$\and
Qianqian Qiao$^1$\and
Heng Huang$^1$\and
Huaye Wang$^1$\\
Yihang Bo$^2$\and
Bao Peng$^2$\and
Jingdong Chen$^{1}$\and
Jun Zhou$^1$\thanks{Corresponding author.}\and
Xin Jin$^{3}$\footnotemark[1]\\
\affiliations
$^1$Ant Group, Beijing, China\\
$^2$Beijing Film Academy, Beijing, China\\
$^3$State Key Laboratory of General Artificial Intelligence, Beijing Institute for General Artificial Intelligence (BIGAI), Beijing, China\\
\emails
\{jianglongteng.jlt, yuandan.zdd, qiaoqianqian.qqq, huangheng.hh, wanghuaye.why\}@antgroup.com, 
boyihang@bfa.edu.cn, 3180100063@zju.edu.cn, jingdongchen.cjd@antgroup.com, 
jun.zhoujun@antfin.com, jinxin@bigai.ai
}

\begin{document}

\maketitle

\begin{figure*}[t]
    \centering
    \includegraphics[width=\textwidth]{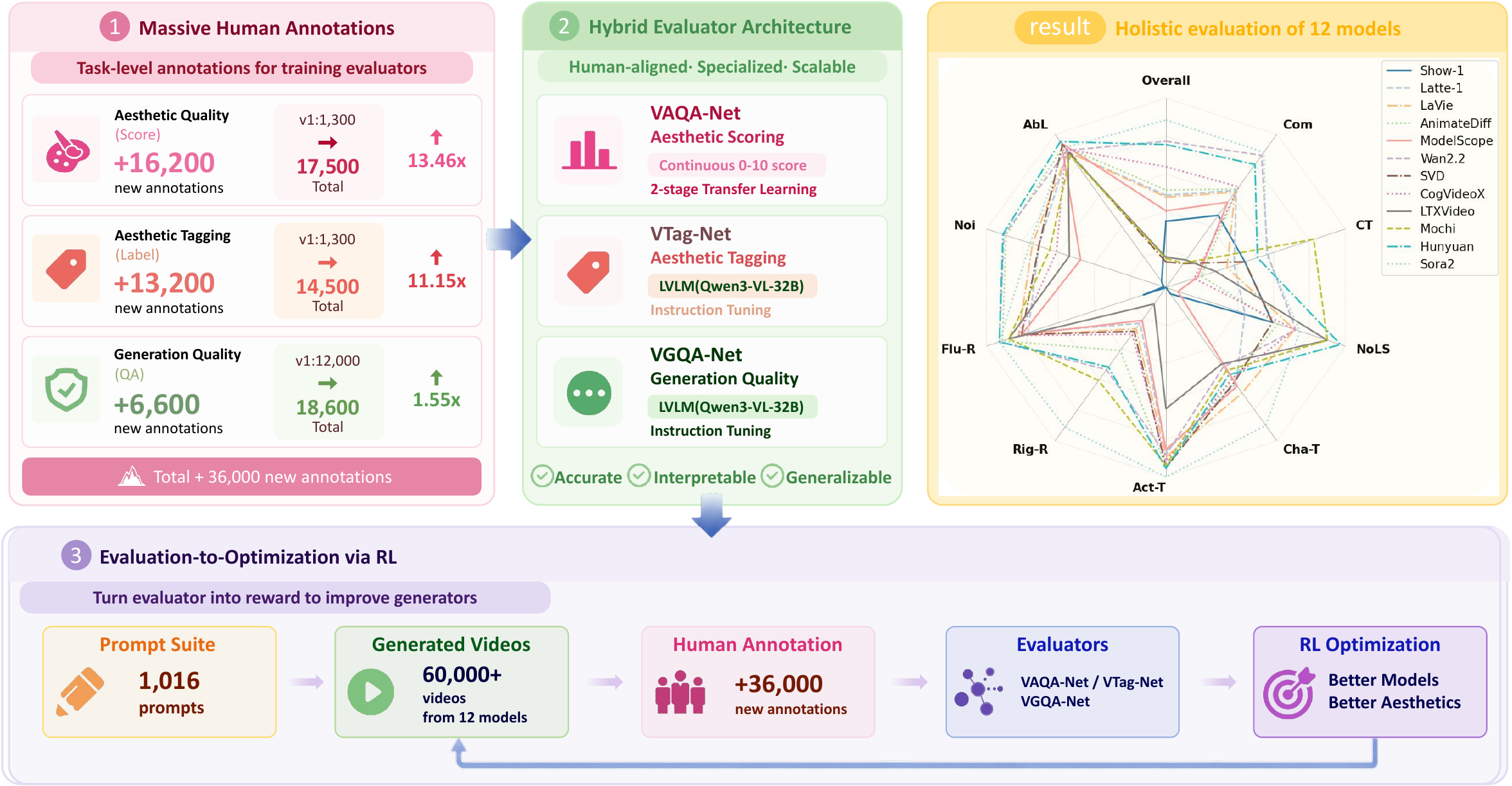}
    \captionof{figure}{Overview of VGA-BenchV2. Left: expanded human annotations supporting evaluator training; middle: hybrid evaluator architecture combining VAQA-Net, VTag-Net, and VGQA-Net; right: holistic evaluation of 12 state-of-the-art video generation models, visualizing performance across 10 representative dimensions selected from the 52 sub-dimensions of VGA-Bench. The bottom panel illustrates the evaluation-to-optimization pipeline using RL.}
    \label{fig:head}
\end{figure*}

\begin{abstract}
The rapid advancement of AIGC video generation calls for evaluation frameworks that move beyond technical fidelity and incorporate human-centered aesthetic assessment. Existing benchmarks often overlook fine-grained perceptual qualities such as visual aesthetics, artistic style, and human preference. To address this limitation, we introduce VGA-BenchV2, an extended human-aligned benchmark and optimization framework for jointly evaluating and improving video generation quality and aesthetic value.

Built upon VGA-Bench, VGA-BenchV2 preserves the original fine-grained taxonomy with two primary dimensions—Aesthetic and Generation—and 52 sub-dimensions. Guided by this taxonomy, we curate 1,016 diverse prompts and collect over 60,000 videos generated by 12 mainstream video generation models. More importantly, VGA-BenchV2 substantially expands human-labeled supervision by adding 36,000 task-level annotations, including 16,200 for aesthetic quality, 13,200 for aesthetic tagging, and 6,600 for generation quality, corresponding to 13.46×, 11.15×, and 1.55× scale-ups over VGA-Bench, respectively.

Leveraging this enlarged annotation corpus, we develop a hybrid evaluator architecture consisting of VAQA-Net for continuous aesthetic scoring and two Qwen-based Large Vision-Language Model evaluators, VTag-Net and VGQA-Net, for aesthetic tagging and generation quality assessment. Extensive experiments demonstrate strong alignment with human judgments across diverse generation models. Beyond evaluation, VGA-BenchV2 further introduces an evaluation-to-optimization pipeline, where the learned aesthetic evaluator serves as a reward model for reinforcement learning-based generator fine-tuning. This closes the loop from benchmark construction and human supervision to automated evaluation and model optimization, enabling video generators to improve not only in realism but also in aesthetic quality and human preference alignment. Resources are available at \url{https://huggingface.co/datasets/BestiVictoryLab/VGA-Bench}.
\end{abstract}

\section{Introduction}
\label{sec:intro}


Text-to-video generation has rapidly evolved from early proof-of-concept systems into increasingly capable generative models that can synthesize coherent, temporally stable, and visually compelling videos from natural language prompts~\cite{blattmann2023stable,blattmann2023align,luo2023videofusion,khachatryan2023text2video}. Recent advances in diffusion models~\cite{blattmann2023stable,song2020score}, transformer-based architectures~\cite{liu2022video,selva2023video}, and large-scale vision-language pretraining~\cite{chen2023vlp,wang2023image,dou2022coarse} have further accelerated this progress, enabling state-of-the-art systems~\cite{wan2025wan,kong2024hunyuanvideo,tang2025human,liu2024sora,hacohen2024ltx,genmo2024mochi,ma2024latte,yang2024cogvideox,wang2023modelscope,zhang2025show,wang2025lavie,guo2023animatediff,blattmann2023stable} to support increasingly diverse creative scenarios. As these models move closer to real-world use in digital art, film production, advertising, and virtual reality, evaluation is no longer limited to verifying whether a video is technically plausible. It must also measure whether the generated content is aesthetically appealing, stylistically controllable, and aligned with human visual preferences.

However, most existing evaluation protocols are still designed around technical correctness. Metrics such as FVD~\cite{unterthiner2019fvd} and CLIP Score~\cite{hessel2021clipscore}, together with their advanced variants~\cite{liu2023fetv}, primarily quantify distributional similarity, temporal consistency, prompt alignment, or low-level visual artifacts. These measurements are useful for diagnosing basic generation quality, but they provide limited insight into perceptual and artistic factors such as composition, lighting, color harmony, cinematic expression, and style controllability. More importantly, conventional metrics usually serve as passive evaluation tools: they can rank or compare models, but are less effective at providing human-aligned supervision that can be directly used to improve video generators.

A series of video generation benchmarks have been proposed to address this evaluation gap. V-Bench~\cite{huang2024vbench} is a representative effort toward standardized multi-dimensional evaluation, but its treatment of video aesthetics remains coarse and depends heavily on off-the-shelf scoring models such as MUSIQ~\cite{ke2021musiq} and DINO~\cite{caron2021emerging}. VGA-Bench~\cite{jiang2026vga} further advances this direction by introducing a fine-grained taxonomy for jointly evaluating aesthetic quality, aesthetic tags, and generation quality. Nevertheless, as video generation models become stronger and more widely used, a benchmark alone is insufficient: reliable aesthetic evaluation requires substantially larger human-labeled supervision, more expressive evaluator architectures, and a mechanism to convert human preference judgments into actionable optimization signals. These requirements motivate VGA-BenchV2, which extends VGA-Bench from a fine-grained evaluation benchmark into a human-aligned framework that unifies large-scale annotation, hybrid automated evaluation, and reinforcement learning-based model optimization.

Notably, V-Bench~\cite{huang2024vbench} represents a pioneering systematic effort to evaluate AIGC videos across multiple dimensions, marking a significant step toward standardized evaluation. However, it reduces the multifaceted nature of video aesthetics into a limited set of scalar measurements and relies heavily on off-the-shelf scoring models, such as MUSIQ~\cite{ke2021musiq} and DINO~\cite{caron2021emerging}. This reliance inevitably leads to coarse granularity, potential domain bias, and limited interpretability for improving generation models. VGA-Bench~\cite{jiang2026vga} further advances this direction by introducing a fine-grained benchmark for jointly evaluating video aesthetic quality, aesthetic tags, and generation quality. Nevertheless, its evaluator training is still limited by the scale of human-labeled supervision, and the benchmark mainly focuses on evaluation rather than forming a closed loop from human preference modeling to generative model optimization.

To address these limitations, this paper introduces VGA-BenchV2 (as shown in Figure~\ref{fig:head}), an extended human-aligned benchmark and optimization framework for video aesthetic and generation quality assessment. Built upon VGA-Bench~\cite{jiang2026vga}, VGA-BenchV2 preserves the original fine-grained taxonomy with two primary dimensions, Aesthetic and Generation, and 52 sub-dimensions, while substantially extending the human annotation scale, evaluator architecture, and optimization capability. Our main contributions are summarized as follows:

\begin{itemize}

\item \textbf{Large-scale human annotation expansion.}
Compared with VGA-Bench~\cite{jiang2026vga}, VGA-BenchV2 substantially expands human-labeled supervision by adding 36,000 newly collected task-level annotations, including 16,200 annotations for aesthetic quality assessment, 13,200 annotations for aesthetic tagging, and 6,600 annotations for generation quality assessment. This corresponds to 13.46$\times$, 11.15$\times$, and 1.55$\times$ scale-ups over VGA-Bench for the three tasks, respectively, transforming the benchmark from an evaluation-oriented suite into a large-scale human-aligned training infrastructure.

\item \textbf{Human-supervised hybrid evaluator architecture.}
Leveraging the enlarged annotation corpus, we develop a hybrid evaluator system consisting of VAQA-Net for continuous aesthetic scoring and two Qwen-based Large Vision-Language Model evaluators, VTag-Net and VGQA-Net, for aesthetic tagging and generation quality assessment, respectively. This design combines specialized aesthetic regression with LVLM-based semantic reasoning, enabling scalable, interpretable, and human-aligned evaluation across diverse generation models.

\item \textbf{Evaluation-to-optimization via reinforcement learning.}
Beyond passive evaluation, VGA-BenchV2 introduces an evaluation-to-optimization pipeline in which the learned aesthetic evaluator serves as a reward model for reinforcement learning-based fine-tuning of video generators. By using the aesthetic score as a reward signal, the framework converts human aesthetic preferences into actionable optimization objectives, demonstrating that VGA-BenchV2 can directly guide generative models toward higher aesthetic quality and better human preference alignment.

\item \textbf{Unified benchmark, evaluator, and optimization infrastructure.}
VGA-BenchV2 integrates 52 fine-grained evaluation dimensions, 1,016 diverse prompts, over 60,000 videos generated by 12 mainstream models, large-scale human annotations, hybrid automated evaluators, and reinforcement learning-based optimization into a unified framework. This enables systematic analysis, fair cross-model comparison, and closed-loop improvement of video generation models.

\end{itemize}

We believe that VGA-BenchV2 serves not only as a rigorous evaluation platform, but also as a key infrastructure for advancing the next generation of video generation systems with enhanced aesthetic intelligence, artistic controllability, and human preference alignment.

\begin{figure}[t]
  \includegraphics[width=\linewidth]{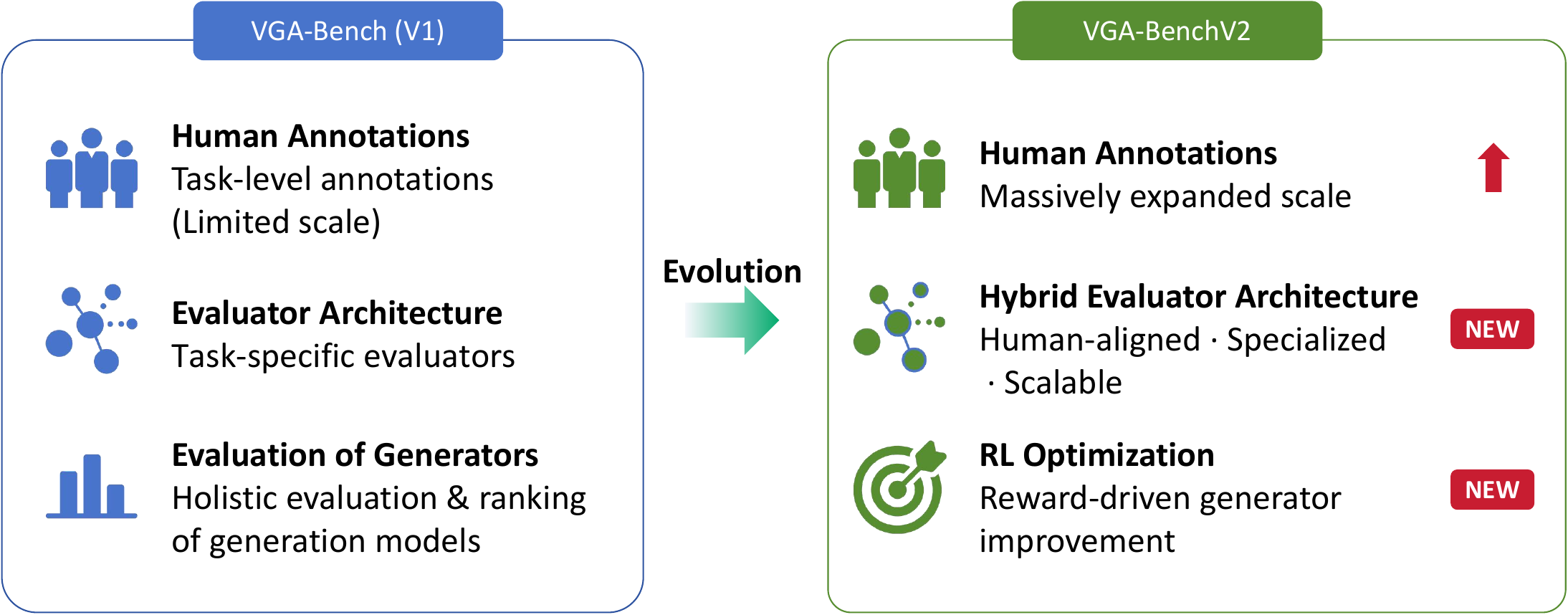}
  \caption{Comparison between VGA-Bench (V1) and VGA-BenchV2 (V2). Both include human annotations and multi-model evaluation; V2 expands annotation scale, adds a hybrid evaluator, and supports optional RL optimization.}
  \label{fig:compare}
\end{figure}

\section{Related Work}
\label{sec:related}

\begin{table*}[t]
    \centering
    \renewcommand{\arraystretch}{1.15}
    \setlength{\tabcolsep}{5pt}
    \begin{tabular}{l c c c c c}
    \hline
    \textbf{Benchmark} 
    & \textbf{\makecell{Aesthetic\\Dims}} 
    & \textbf{\makecell{Evaluator\\Architecture}} 
    & \textbf{\makecell{Annotation\\Usage}} 
    & \textbf{\makecell{RL\\Opt.}} \\
    \hline
    V-Bench~\cite{huang2024vbench} 
     & 1 & Off-the-shelf Expert Models & Alignment Only & \ding{55} \\
    
    V-Bench2.0~\cite{zheng2025vbench} 
     & 2 & Off-the-shelf Expert Models & Alignment Only & \ding{55} \\
    
    T2V-CompBench~\cite{sun2025t2v} 
     & 0 & MLLM-based & Alignment Only & \ding{55} \\
    
    ChronoMagic-Bench~\cite{yuan2024chronomagic} 
     & 0 & Hybrid & Alignment Only & \ding{55} \\
    
    StoryEval~\cite{wang2025your} 
     & 0 & MLLM-based & Alignment Only & \ding{55} \\
    
    VGA-Bench~\cite{jiang2026vga} 
     & \textbf{21} & Specialized Nets & Training \& Alignment & \ding{55} \\
    \hline
    \textbf{VGA-BenchV2 (Ours)} 
    & \textbf{21} 
    & \textbf{\makecell{Hybrid\\Specialized Nets + LVLMs}} 
    & \textbf{\makecell{Training, Alignment\\\& Reward}} 
    & \textbf{\ding{51}} \\
    \hline
    \end{tabular}
    \caption{Comparison with existing video generation benchmarks. VGA-BenchV2 preserves the fine-grained aesthetic and generation taxonomy of VGA-Bench while further extending it with hybrid human-aligned evaluators and an evaluation-to-optimization pipeline.}
    \label{tab:comparison}
\end{table*}





Evaluating video generation models remains challenging despite their rapid progress. Early metrics such as FVD~\cite{unterthiner2019fvd}, CLIP Score~\cite{hessel2021clipscore}, and their advanced variants~\cite{liu2023fetv} mainly measure distributional similarity, temporal consistency, prompt alignment, or low-level visual artifacts. While useful for assessing basic technical quality, these metrics are insufficient for capturing fine-grained perceptual factors such as composition, lighting, color harmony, artistic style, and overall aesthetic appeal.

Recent benchmarks have promoted more systematic evaluation of video generation models. V-Bench~\cite{huang2024vbench} provides a representative multi-dimensional evaluation protocol, and V-Bench2~\cite{zheng2025vbench} further extends the evaluation scope. Other benchmarks, including ChronoMagic-Bench~\cite{yuan2024chronomagic}, T2V-CompBench~\cite{sun2025t2v}, and StoryEval~\cite{wang2025your}, focus on complementary aspects such as temporal coherence, compositional binding, and narrative consistency. Compared with these works, VGA-Bench~\cite{jiang2026vga} is distinguished by its fine-grained treatment of video aesthetics, jointly modeling aesthetic quality, aesthetic tags, and generation quality within a unified taxonomy.

VGA-BenchV2 further extends VGA-Bench~\cite{jiang2026vga} from a fine-grained evaluation benchmark into a human-aligned evaluation and optimization framework. Specifically, it strengthens human-labeled supervision, introduces a hybrid evaluator architecture that combines specialized aesthetic scoring with LVLM-based tagging and generation-quality assessment, and further connects evaluation with reinforcement learning-based generator optimization. In this way, VGA-BenchV2 moves beyond passive benchmarking and forms a closed-loop framework that links benchmark design, human annotation, automated evaluation, and model improvement.

\section{VGA-BenchV2 Construction}
\label{sec:Suite}

\subsection{Inherited Evaluation Taxonomy}

VGA-BenchV2 inherits the fine-grained evaluation taxonomy of VGA-Bench~\cite{jiang2026vga} to ensure continuity, reproducibility, and cross-version comparability. Rather than redesigning the benchmark dimensions, we preserve the original taxonomy as the evaluation backbone and focus the V2 extension on human annotation expansion, hybrid evaluator training, and evaluation-to-optimization. Overall, the taxonomy contains 52 sub-dimensions organized into two primary categories: \textbf{Aesthetic} and \textbf{Generation}.

The \textbf{Aesthetic} category consists of two complementary perspectives. 

\begin{itemize}
\item \textbf{Aesthetic Quality} measures the perceptual appeal of generated videos as continuous scores, covering ten dimensions adapted from VADB~\cite{qiao2025vadb}: Overall Score (Ovr), Composition (Com), Shot Size (SS), Lighting (Lig), Visual Tone (VT), Color (Col), Depth of Field (DoF), Expression (Exp), Costume (Cos), and Makeup (Mak). 

\item \textbf{Aesthetic Tagging} describes discrete visual and stylistic attributes as classification labels. Following photographic and cinematographic principles~\cite{qiao2025vadb,matbouly2022quantifying,deren1960cinematography,brown2016cinematography}, it contains eleven tag dimensions: Composition Types (CT), Shot Type (ST), Number of Light Sources (NoLS), Light Source Position (LSP), Light Quality (LQ), Light Color (LC), Color Temperature (ColT), Saturation (Sat), Brightness (Bri), Contrast (Con), and Depth of Field (DoF). These 21 aesthetic-related dimensions support both quantitative model ranking and interpretable analysis of visual style controllability.
\end{itemize}

The \textbf{Generation} category evaluates whether generated videos are semantically faithful, physically plausible, and visually stable. Following VGA-Bench~\cite{jiang2026vga} and building upon V-Bench~\cite{huang2024vbench}, VGA-BenchV2 retains 31 generation-quality sub-dimensions grouped into three aspects. 

\begin{itemize}
\item \textbf{Video-Text Consistency} includes Character-Text Consistency (Cha-T), Action-Text Consistency (Act-T), Object-Text Consistency (Obj-T), Scene-Text Consistency (Sce-T), Object Position-Text Consistency (Pos-T), Camera Movement-Text Consistency (Cam-T), Object Attribute-Text Consistency (Att-T), Video Content-Text Consistency (Cnt-T), Video Style-Text Consistency (Sty-T), and Video Speed-Text Consistency (Spd-T). 

\item \textbf{Reality \& Plausibility} includes Rigid Body Collision Realism (Rig-R), Fluid Motion Realism (Flu-R), Gaseous Motion Realism (Gas-R), Gradual Change Realism (Gra-R), Object Trajectory Realism (Tra-R), Action Realism (Act-R), Scene Realism (Sce-R), Object Realism (Obj-R), Character Generation Quality (Cha-R), Weather Representation Realism (Wea-R), Time Period Representation Realism (Tim-R), Lighting \& Shadow Realism (Lig-R), Moving Scene Reasonableness (Dyn-R), Textual Attribute Representation (Att-R), Overall Realism (Glo-R), and Unrealistic Description Imaginative Presentation (Ima-R). 

\item \textbf{Basic Quality} includes Video Noise-Free (Noi), Abnormal Lighting Detection (AbL), Video Clarity (Cla), Static Content Stability (Sta), and Static Content Non-distortion (Dis). We use the suffix ``\textbf{-T}'' to denote video-text consistency dimensions and ``\textbf{-R}'' to denote reality or plausibility dimensions, while basic quality dimensions use standalone mnemonic codes. Detailed definitions and evaluation criteria are provided in the Appendix.
\end{itemize}

\begin{figure}[t]
  \includegraphics[width=\linewidth]{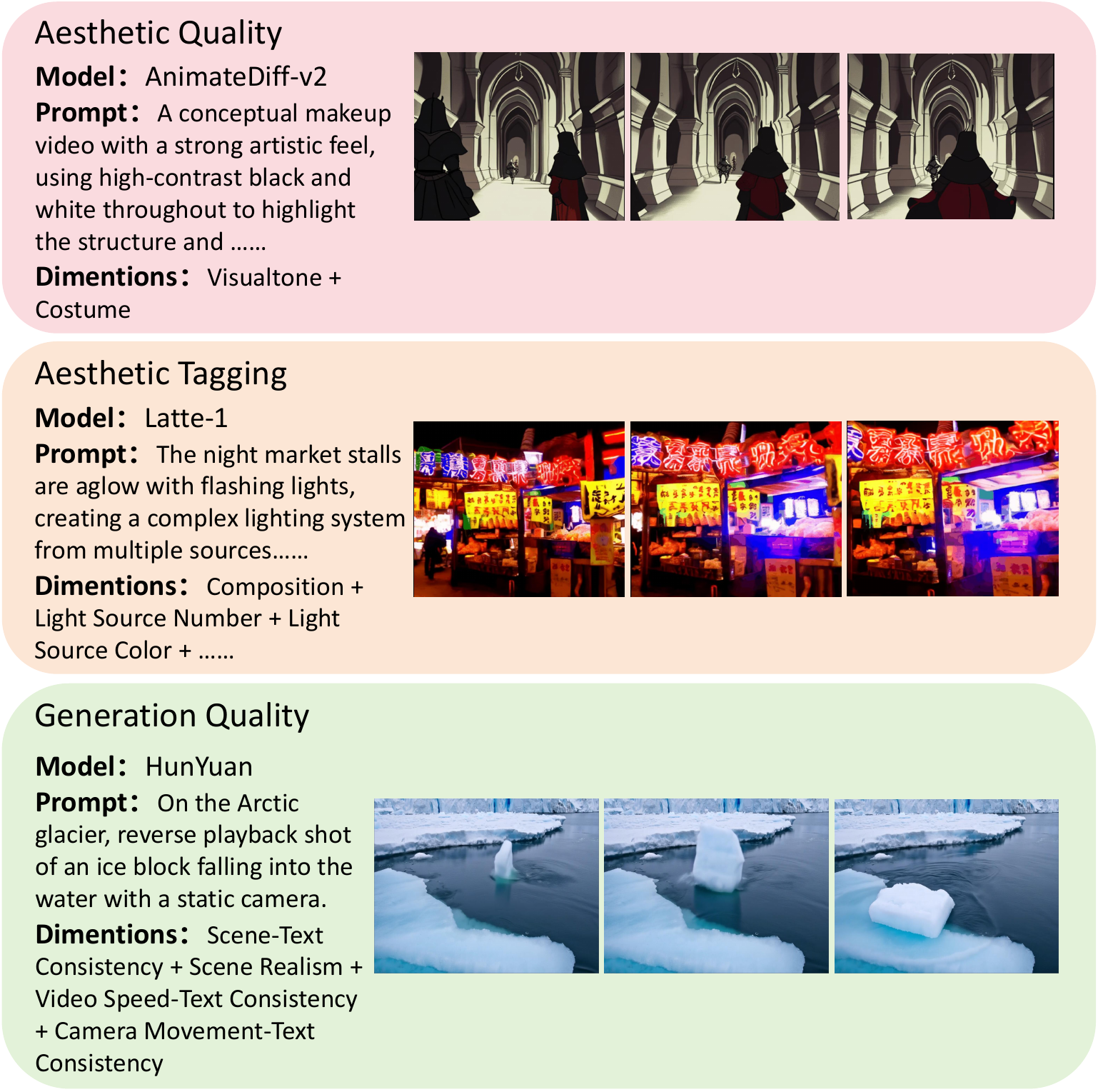}
  \caption{Prompts for the three core dimensions, their corresponding sub-dimensions, and example generated videos.}
  \label{fig:sample-A}
\end{figure}

\subsection{Prompt Suite and Video Pool}
VGA-BenchV2 inherits the prompt suite from VGA-Bench~\cite{jiang2026vga} to ensure cross-version comparability and stable dimension-wise evaluation. The prompt suite is designed under the principle of explicit attribute specification: each prompt explicitly triggers one or more target dimensions, such that annotators and automated evaluators assess only the attributes that are semantically grounded in the prompt. This design reduces ambiguity during both human annotation and automated evaluation.

The prompt suite contains 1,016 prompts, including 200 prompts for aesthetic quality, 220 for aesthetic tagging, and 596 for generation quality. Each dimension is covered by at least 50 prompts, and the suite supports both single-dimension and multi-dimensional combinations. We also retain two lightweight subsets with 508 and 127 prompts for efficient evaluation under different computational budgets.

In VGA-BenchV2, the prompt suite serves as a unified anchor for three downstream components: large-scale human annotation, supervised evaluator training, and evaluation-to-optimization experiments. By reusing the same dimension-aligned prompts as VGA-Bench, VGA-BenchV2 isolates the effect of expanded human supervision and improved evaluator design, while preserving a consistent benchmark protocol for comparing video generation models.

\begin{figure}[t]
  \includegraphics[width=\linewidth]{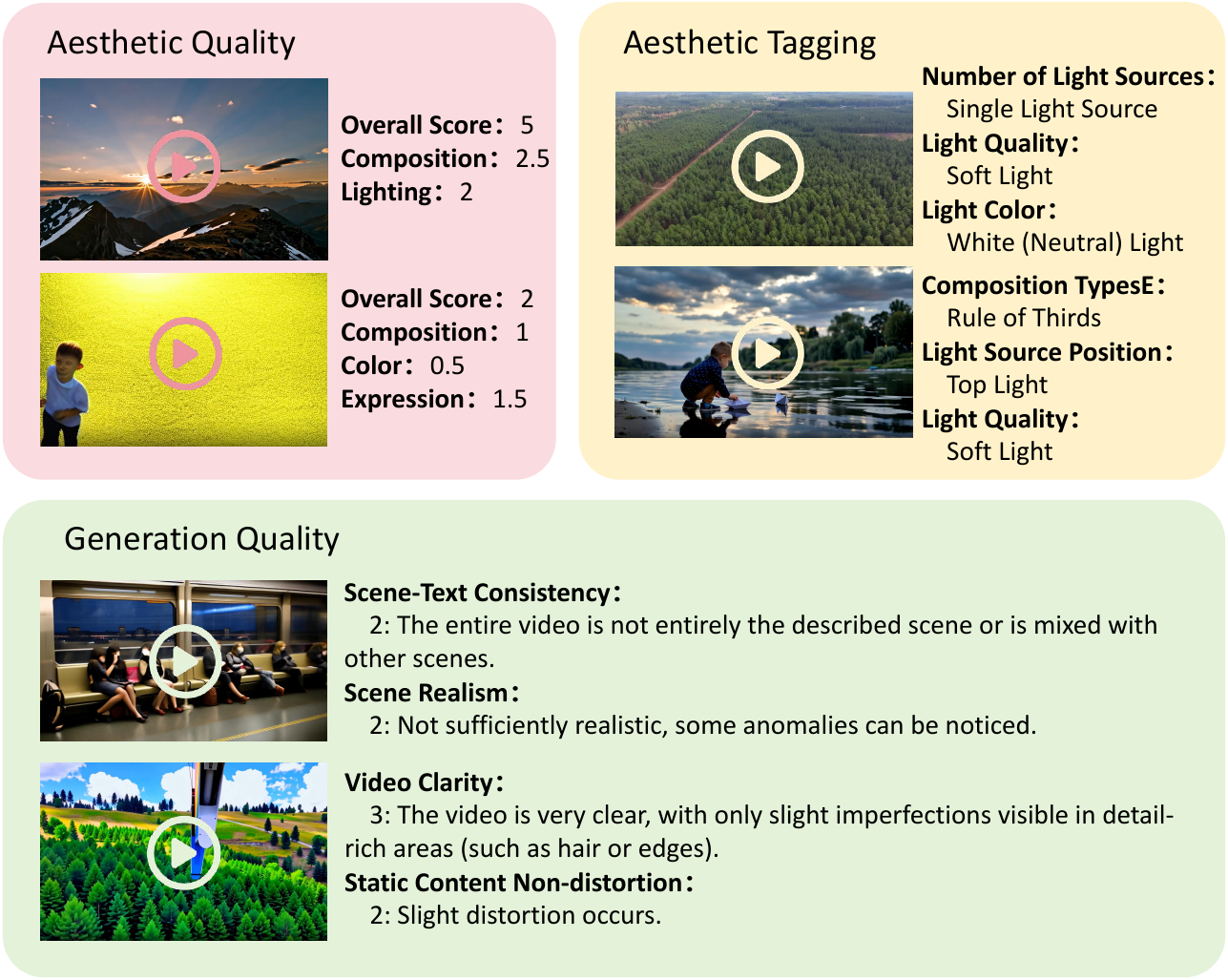}
  \caption{Examples of human annotations for the three core dimensions.}
  \label{fig:sample-B}
\end{figure}

\subsection{Large-scale Human Annotation Expansion}

A central goal of VGA-BenchV2 is to transform VGA-Bench from an evaluation-oriented benchmark into a human-aligned training infrastructure for automated video evaluators. To this end, we substantially expand the task-level human annotations over VGA-Bench~\cite{jiang2026vga}. Specifically, VGA-BenchV2 introduces 36,000 newly collected annotations, including 16,200 annotations for aesthetic quality assessment, 13,200 annotations for aesthetic tagging, and 6,600 annotations for generation quality assessment. These annotations provide direct supervision for evaluator training, alignment verification, and reward modeling.

\begin{table}[t]
    \centering
    \scriptsize
    \renewcommand{\arraystretch}{1.15}
    \setlength{\tabcolsep}{4pt}
    \begin{tabular}{l c c c c}
    \hline
    \textbf{Task} 
    & \textbf{\makecell{VGA-Bench\\Supervision}} 
    & \textbf{\makecell{New V2\\Annotations}} 
    & \textbf{\makecell{V2\\Total}} 
    & \textbf{Scale-up} \\
    \hline
    Aesthetic Quality 
    & 1,300 & +16,200 & 17,500 & 13.46$\times$ \\
    Aesthetic Tagging 
    & 1,300 & +13,200 & 14,500 & 11.15$\times$ \\
    Generation Quality 
    & 12,000 & +6,600 & 18,600 & 1.55$\times$ \\
    \hline
    \textbf{Total} 
    & \textbf{14,600} & \textbf{+36,000} & \textbf{50,600} & \textbf{3.47$\times$} \\
    \hline
    \end{tabular}
    \caption{Human annotation expansion in VGA-BenchV2. We report task-level supervision for aesthetic quality, aesthetic tagging, and generation quality. VGA-BenchV2 substantially increases the amount of human-labeled data used for evaluator training and alignment.}
    \label{tab:annotation_expansion}
\end{table}

\subsubsection{Annotation Protocol and Quality Control}

We follow an expert-guided and multi-annotator annotation protocol. Domain experts first provide exemplar annotations and detailed rating guidelines for each task. Trained annotators then label the remaining samples according to these exemplars, while experts conduct batch-wise audits to ensure annotation consistency. If a batch fails the quality check, it is rejected and re-annotated.

For aesthetic quality, each sample is scored on a 0--10 scale, and the final score is obtained by averaging multiple independent ratings. For aesthetic tagging, annotators assign discrete visual labels such as composition type, light source position, shot type, color temperature, and contrast, with final labels determined by majority voting. For generation quality, each dimension is formulated as a task-specific question with structured ordinal options, enabling the labels to capture different levels of semantic consistency, realism, and basic visual quality.

Importantly, all annotation tasks follow the explicit-trigger principle inherited from the prompt suite: annotators evaluate only the dimensions explicitly specified by the prompt. This prevents subjective speculation on unmentioned attributes and ensures that each label is semantically grounded in the prompt-video pair.

\subsection{Hybrid Automated Evaluation Networks}

Leveraging the expanded human-annotated corpus, VGA-BenchV2 moves beyond lightweight task-specific evaluators to train a robust, human-aligned hybrid evaluator system. The architecture combines specialized aesthetic regression with LVLM-based semantic reasoning, consisting of VAQA-Net for continuous aesthetic scoring, VTag-Net for aesthetic tagging, and VGQA-Net for generation quality assessment.

\textbf{VAQA-Net (Aesthetic Scoring):} We adopt a two-stage transfer learning strategy. First, the video encoder is initialized from the VADB pre-trained model to inherit aesthetic representations. Second, the model is fine-tuned on a large-scale generated video dataset, including 8,366 videos from 12 mainstream generation models with high-quality human annotations. Evaluation is performed on a held-out set of 1,186 generated videos to assess generalization.

\textbf{VTag-Net \& VGQA-Net (LVLM-based Evaluators):} To enable semantic reasoning and detailed assessment, we leverage Qwen3-VL-32B~\cite{Qwen3-VL} as the backbone and apply instruction tuning for task-specific adaptation. 
\begin{itemize} 
\item \textbf{VTag-Net:} Trained on a combination of real VADB videos and 11,100 generated videos with human-annotated aesthetic tags, outputting discrete visual labels.
\item \textbf{VGQA-Net:} Trained on 7,054 generated videos with human-annotated question-answer pairs, optimized for coherent reasoning and generation quality assessment.
\end{itemize} 
The training set includes representative samples from all 12 source models to ensure generalization across generation paradigms.

\begin{figure}[t]
  \centering
  \includegraphics[width=\linewidth]{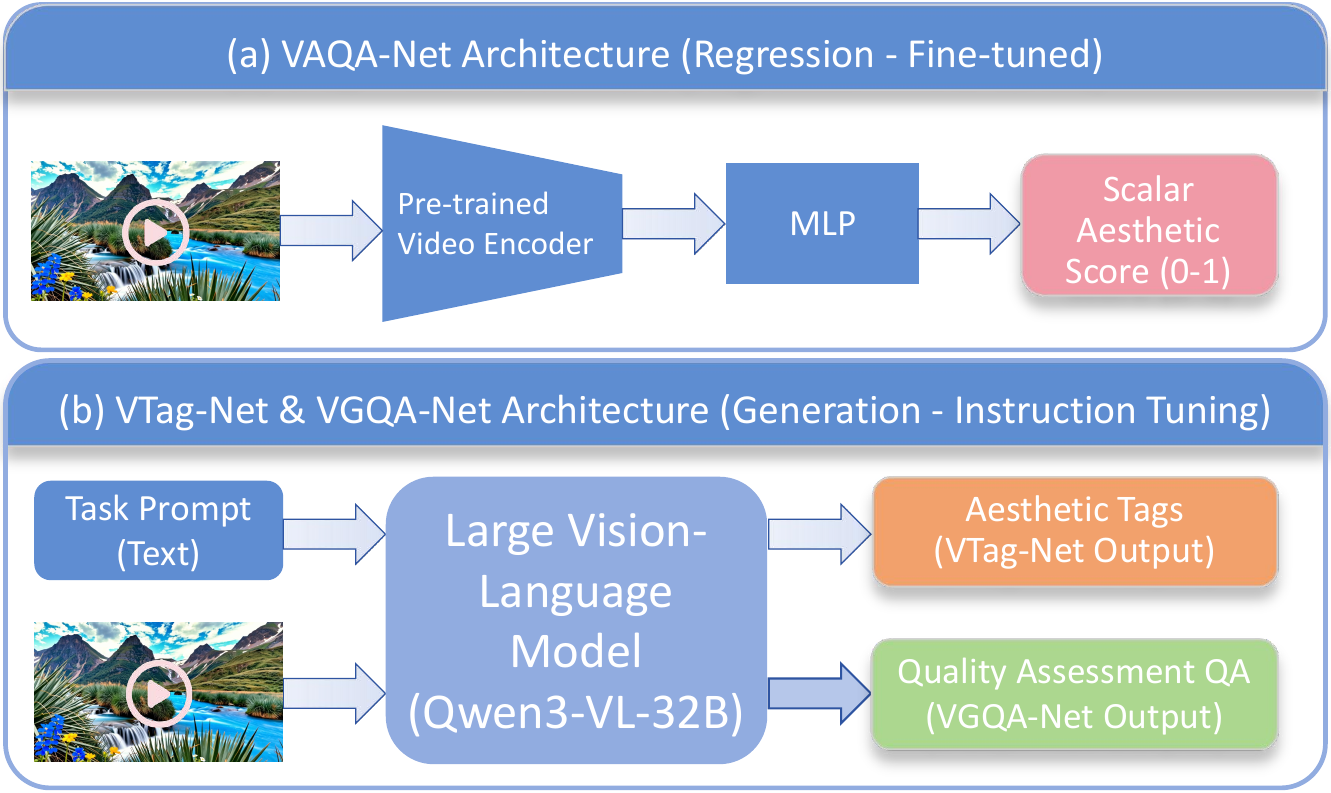}
  \caption{Overview of the hybrid evaluation networks. VAQA-Net predicts scalar aesthetic scores via a pre-trained encoder and regression head, while VTag-Net and VGQA-Net leverage a LVLM backbone via instruction tuning for aesthetic tagging and quality reasoning.}
  \label{fig:model}
\end{figure}

The network architectures of VAQA-Net, VTag-Net, and VGQA-Net are illustrated in Figure~\ref{fig:model}.

\subsection{Evaluation-to-Optimization Interface}
\label{sec:EvalToOpt}

Beyond serving as passive evaluation metrics, the learned evaluators in VGA-BenchV2 can be used as reward models for generator optimization. In particular, the aesthetic score predicted by VAQA-Net provides a scalar reward reflecting human-aligned visual preference. This reward can be incorporated into reinforcement learning-based fine-tuning pipelines, enabling video generators to optimize toward higher aesthetic quality while maintaining stability.

Formally, given a prompt $p$, the current policy $\pi_{\theta}$ generates a video $v \sim \pi_{\theta}(\cdot|p)$, and VAQA-Net outputs a scalar reward:
\begin{equation}
r_{\phi}(v) = \mathrm{VAQA}_{\phi}(v),
\end{equation}
with $\phi$ denoting the frozen evaluator parameters. To stabilize training, KL regularization is applied:
\begin{equation}
\tilde{r}(v,p) = r_{\phi}(v) - \lambda_{\mathrm{KL}} 
D_{\mathrm{KL}}\left(\pi_{\theta}(\cdot|p)\,\|\,\pi_{\mathrm{ref}}(\cdot|p)\right),
\end{equation}
and the normalized advantage is
\begin{equation}
A_i = \frac{\tilde{r}_i - \mu_{\mathcal{B}}}{\sigma_{\mathcal{B}}+\epsilon},
\end{equation}
where $\mu_{\mathcal{B}}$ and $\sigma_{\mathcal{B}}$ denote the batch mean and standard deviation.

For policy optimization, the unclipped and clipped advantages are defined as
\begin{equation}
\begin{split}
r_{i,t} &= \rho_{i,t} A_i, \\
c_{i,t} &= \mathrm{clip}(\rho_{i,t}, 1-\epsilon, 1+\epsilon) A_i, \\
\rho_{i,t} &= \frac{\pi_{\theta}(a_{i,t}|s_{i,t})}{\pi_{\theta_{\mathrm{old}}}(a_{i,t}|s_{i,t})},
\end{split}
\end{equation}
and the clipped GRPO objective is
\begin{equation}
\mathcal{L}_{\mathrm{GRPO}}(\theta) = - \mathbb{E}_{i,t} \big[\min(r_{i,t}, c_{i,t}) \big].
\end{equation}
This formulation encourages higher-reward samples to contribute stronger updates while constraining policy drift. Detailed implementation and training parameters are described in Section~\ref{sec:Experiments}.

\section{Experiments and Results}
\label{sec:Experiments}

\subsection{Validation of Automated Evaluator}

\begin{table}[t]
    \centering
    \footnotesize
    \setlength{\tabcolsep}{3pt} 
    \renewcommand{\arraystretch}{1.1} 
    \begin{tabular}{lc|lcc}
        \toprule
        \multicolumn{2}{c|}{\textbf{VAQA-Net (SROCC)}} & \multicolumn{3}{c}{\textbf{VTag-Net (Accuracy)}} \\
        \midrule
        \textbf{Dimension} & \textbf{Score} & \textbf{Dimension} & \textbf{\#Cls.} & \textbf{Acc.} \\
        \midrule
        \textbf{Overall Score} & \textbf{87.6} & Color Temp. (ColT) & 3 & 64.8 \\
        Composition (Com) & 86.9 & Saturation (Sat) & 3 & 71.9 \\
        Shot Size (SS) & 87.4 & Brightness (Bri) & 3 & 70.3 \\
        Lighting (Lig) & 87.2 & Contrast (Con) & 3 & 75.0 \\
        Visual Tone (VT) & 87.4 & Light Quality (LQ) & 2 & 66.3 \\
        Color (Col) & 86.5 & Light Color (LC) & 4 & 93.6 \\
        Depth of Field (DoF) & 87.4 & Light Pos. (LSP) & 5 & 68.3* \\
        Expression (Exp) & 89.2 & Num. Lights (NoLS) & 2 & 83.7 \\
        Costume (Cos) & 88.4 & Comp. Type (CT) & 7 & 49.8* \\
        Makeup (Mak) & 86.3 & Shot Type (ST) & 5 & 75.9* \\
         & & Depth of Field (DoF) & 2 & 76.1 \\
        \bottomrule
    \end{tabular}
    \caption{Validation of Aesthetic Evaluators. We report the SROCC (\%) for VAQA-Net and Classification Accuracy (\%) for VTag-Net. Specifically, we adopt Top-1 Accuracy for dimensions with $\le 4$ classes and Top-2 Accuracy for those with $> 4$ classes (marked with *) to account for varying label space sizes.}
    \label{tab:aesthetic_val}
\end{table}

\begin{table}[t]
    \centering
    \setlength{\tabcolsep}{5pt} 
    \renewcommand{\arraystretch}{1.1} 
    \resizebox{\linewidth}{!}{
        \begin{tabular}{lc lc lc lc}
            \toprule
            \textbf{Dim.} & \textbf{Acc.} & \textbf{Dim.} & \textbf{Acc.} & \textbf{Dim.} & \textbf{Acc.} & \textbf{Dim.} & \textbf{Acc.} \\
            \midrule
            Cha-T & 70.5 & Sty-T & 69.6 & Sce-R & 89.2 & Glo-R & 69.8 \\
            Act-T & 70.7 & Spd-T & 76.6 & Obj-R & 81.2 & Ima-R & 59.2 \\
            Obj-T & 75.2 & Rig-R & 62.2 & Cha-R & 56.8 & Noi & 58.3 \\
            Sce-T & 74.7 & Flu-R & 75.6 & Wea-R & 66.8 & AbL & 85.7 \\
            Pos-T & 64.3 & Gas-R & 78.3 & Tim-R & 83.1 & Cla & 66.4 \\
            Cam-T & 72.5 & Gra-R & 80.4 & Lig-R & 80.5 & Sta & 50.5 \\
            Att-T & 72.6 & Tra-R & 62.1 & Dyn-R & 67.0 & Dis & 71.7 \\
            Cnt-T & 58.4 & Act-R & 80.2 & Att-R & 78.7 & -- & -- \\
            \bottomrule
        \end{tabular}
    }
    \caption{Validation of VGQA-Net. We report the Accuracy (\%) across 31 sub-dimensions. The suffixes -T and -R denote Video-Text Consistency and Reality/Plausibility, respectively.}
    \label{tab:gen_val}
\end{table}

To ensure the reliability of VGA-BenchV2, we rigorously validated our automated evaluators against human ground truth on a held-out test set. For aesthetics, Table \ref{tab:aesthetic_val} shows that VAQA-Net achieves a high SROCC of 87.6\% on the overall score, demonstrating strong ranking consistency with human experts. Simultaneously, VTag-Net delivers robust classification performance, particularly in objective dimensions like Light Color (93.6\%) and Number of Light Sources (83.7\%), validating its precision in identifying visual elements. For generation quality, VGQA-Net (Table \ref{tab:gen_val}) exhibits consistent alignment across all 31 fine-grained dimensions, with an average accuracy approximately 71.3\%. High scores in challenging categories like Scene Realism (89.2\%) and Abnormal Lighting Detection (85.7\%) confirm that our MLLM-based evaluators serve as effective, scalable proxies for human assessment. 

These high scores not only demonstrate the effectiveness of our models but also indicate that their predictions are closely aligned with expert human annotations, corroborating the human-aligned supervision highlighted in Table~\ref{tab:comparison}.

\subsection{VGA-BenchV2 Evaluation Results}
We evaluate all generative models using the trained VAQA-Net, VTag-Net, and VGQA-Net to assess their performance across diverse aesthetic and quality dimensions. To ensure a fair and unbiased ranking, all generated videos used for evaluation are strictly held out from the evaluator training process, preventing data leakage. The evaluated models are listed chronologically by release date: Stable Video Diffusion (SVD)~\cite{blattmann2023stable}, AnimateDiff-v2~\cite{guo2023animatediff}, LaVie~\cite{wang2025lavie}, Show-1~\cite{zhang2025show}, ModelScope~\cite{wang2023modelscope}, CogVideoX~\cite{yang2024cogvideox}, Latte-1~\cite{ma2024latte}, Mochi~\cite{genmo2024mochi}, LTXVideo~\cite{hacohen2024ltx}, HunyuanVideo~\cite{kong2024hunyuanvideo}, Wan2.2~\cite{wan2025wan}, and Sora2~\cite{liu2024sora}.

The ranking metrics are defined as follows: \begin{itemize} \item \textbf{Aesthetic Quality:} Models are ranked by their average predicted score (higher is better). \item \textbf{Aesthetic Tagging:} We utilize alignment accuracy as the metric (i.e., whether the generated video successfully reflects the prompted aesthetic tag), where higher mean accuracy indicates superior controllability. \item \textbf{Generation Quality:} Models are ranked by the mean score across all sub-dimensions, where a higher average indicates better generation fidelity and consistency. \end{itemize} The final aggregated results, obtained by normalizing and averaging scores across all sub-dimensions within the three main categories, are presented in Table \ref{tab:biao6}. For detailed performance breakdowns of each model across specific sub-dimensions, please refer to the Appendix.

\subsection{Aesthetic Optimization via Reinforcement Learning}

\begin{figure}[t]
  \includegraphics[width=\linewidth]{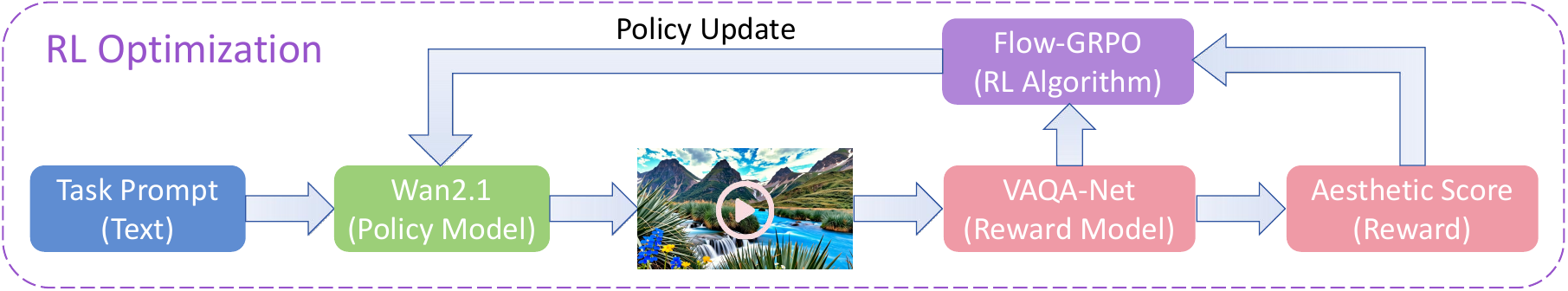}
  \caption{RL Aesthetic Optimization Pipeline. We utilize the Flow-GRPO algorithm to fine-tune video generation models, employing VAQA-Net as the reward model to enhance artistic quality.}
  \label{fig:rl_module}
\end{figure}

To validate the practical effect of our evaluators, we fine-tune Wan2.1 using the evaluation-to-optimization interface described in Section~\ref{sec:EvalToOpt}. As shown in Figure~\ref{fig:rl_module}, the VAQA-Net Overall Score is used as the reward, and Flow-GRPO~\cite{liu2025flow} is applied with LoRA and ODE-to-SDE conversion.

Quantitative results on a held-out test set show that the average aesthetic score increases from 0.49 to 0.52. Qualitative results indicate that fine-tuned videos exhibit more appealing visual composition and higher overall aesthetic quality. These results demonstrate that VGA-BenchV2 evaluators can support not only passive evaluation but also reward-driven video generator optimization.

\begin{table}

    \centering
    \setlength{\tabcolsep}{1.5pt}
    \small 

    \begin{tabular}{l c c c}

    \hline

     Model & Aes. Score & Tag Cla. & Gen. Level \\

    \hline

    SVD~\cite{blattmann2023stable}             & 0.20 & 0.60 & 0.69 \\

    AnimateDiff~\cite{guo2023animatediff}      & 0.36 & 0.57 & 0.67 \\

    LaVie~\cite{wang2025lavie}                 & 0.34 & 0.58 & 0.68 \\

    Show-1~\cite{zhang2025show}                & 0.29 & 0.60 & 0.04 \\

    ModelScope~\cite{wang2023modelscope}       & 0.31 & 0.51 & 0.64 \\

    CogVideoX~\cite{yang2024cogvideox}         & 0.41 & 0.56 & 0.70 \\

    Latte-1~\cite{ma2024latte}                 & 0.35 & 0.60 & 0.68 \\

    Mochi~\cite{genmo2024mochi}                & 0.21 & \textbf{0.68} & 0.71 \\

    LTXVideo~\cite{hacohen2024ltx}             & 0.22 & 0.61 & 0.61 \\

    HunyuanVideo~\cite{kong2024hunyuanvideo}        & 0.45 & \underline{0.66} & \underline{0.73} \\

    Wan2.2~\cite{wan2025wan}                   & \underline{0.46} & 0.63 & 0.71 \\

    Sora2~\cite{liu2024sora}                   & \textbf{0.50} & 0.64 & \textbf{0.80} \\

    \hline

    \end{tabular}

    \caption{Performance comparison of state-of-the-art text-to-video generation models on aesthetic score (Aes.\ Score), tag classification accuracy (Tag Cla.), and generation level (Gen.\ Level) metrics.}

    \label{tab:biao6}

\end{table}

\section{Conclusion}

We present \textbf{VGA-BenchV2}, an extended human-aligned benchmark and optimization framework that builds upon VGA-Bench by preserving 52 sub-dimensions, 1,016 prompts, and over 60,000 videos while introducing three key advancements: expanded human-labeled supervision for aesthetic quality, aesthetic tagging, and generation quality; a hybrid evaluator architecture combining VAQA-Net and LVLM-based VTag-Net and VGQA-Net for accurate, interpretable, and scalable assessment; and an evaluation-to-optimization pipeline enabling RL-based fine-tuning of video generators to align outputs with human aesthetic preference. Experiments demonstrate that VGA-BenchV2 provides more precise, human-aligned evaluation and facilitates multi-dimensional optimization. In the future, it can be extended to higher-resolution videos, additional aesthetic dimensions, cross-modal generation, and controllable creative applications, supporting next-generation AIGC video systems that harmonize fidelity, artistry, and human preference.

\appendix

\section*{Ethical Statement}

We strictly ensure that all prompts and generated videos within this benchmark have been rigorously screened to exclude pornographic, violent, or otherwise offensive content. Furthermore, all human annotation procedures comply with ethical guidelines and have been reviewed and approved by the relevant Institutional Review Board (IRB).

\section*{Acknowledgments}

This work was supported by the Ant Group Research Fund, the National Natural Science Foundation of China under Grant No.62072014, and the Opening Project of the State Key Laboratory of General Artificial Intelligence, BIGAI/Peking University, Beijing, China (Project No.SKLAGI2025OP01).

\bibliographystyle{named}
\bibliography{ijcai26}

@article{blattmann2023stable,
  title={Stable video diffusion: Scaling latent video diffusion models to large datasets},
  author={Blattmann, Andreas and Dockhorn, Tim and Kulal, Sumith and Mendelevitch, Daniel and Kilian, Maciej and Lorenz, Dominik and Levi, Yam and English, Zion and Voleti, Vikram and Letts, Adam and others},
  journal={arXiv preprint arXiv:2311.15127},
  year={2023}
}

@inproceedings{blattmann2023align,
  title={Align your latents: High-resolution video synthesis with latent diffusion models},
  author={Blattmann, Andreas and Rombach, Robin and Ling, Huan and Dockhorn, Tim and Kim, Seung Wook and Fidler, Sanja and Kreis, Karsten},
  booktitle={Proceedings of the IEEE/CVF conference on computer vision and pattern recognition},
  pages={22563--22575},
  year={2023}
}

@article{luo2023videofusion,
  title={Videofusion: Decomposed diffusion models for high-quality video generation},
  author={Luo, Zhengxiong and Chen, Dayou and Zhang, Yingya and Huang, Yan and Wang, Liang and Shen, Yujun and Zhao, Deli and Zhou, Jingren and Tan, Tieniu},
  journal={arXiv preprint arXiv:2303.08320},
  year={2023}
}

@inproceedings{khachatryan2023text2video,
  title={Text2video-zero: Text-to-image diffusion models are zero-shot video generators},
  author={Khachatryan, Levon and Movsisyan, Andranik and Tadevosyan, Vahram and Henschel, Roberto and Wang, Zhangyang and Navasardyan, Shant and Shi, Humphrey},
  booktitle={Proceedings of the IEEE/CVF International Conference on Computer Vision},
  pages={15954--15964},
  year={2023}
}

@article{song2020score,
  title={Score-based generative modeling through stochastic differential equations},
  author={Song, Yang and Sohl-Dickstein, Jascha and Kingma, Diederik P and Kumar, Abhishek and Ermon, Stefano and Poole, Ben},
  journal={arXiv preprint arXiv:2011.13456},
  year={2020}
}

@inproceedings{liu2022video,
  title={Video swin transformer},
  author={Liu, Ze and Ning, Jia and Cao, Yue and Wei, Yixuan and Zhang, Zheng and Lin, Stephen and Hu, Han},
  booktitle={Proceedings of the IEEE/CVF conference on computer vision and pattern recognition},
  pages={3202--3211},
  year={2022}
}

@article{selva2023video,
  title={Video transformers: A survey},
  author={Selva, Javier and Johansen, Anders S and Escalera, Sergio and Nasrollahi, Kamal and Moeslund, Thomas B and Clap{\'e}s, Albert},
  journal={IEEE Transactions on Pattern Analysis and Machine Intelligence},
  volume={45},
  number={11},
  pages={12922--12943},
  year={2023},
  publisher={IEEE}
}

@article{chen2023vlp,
  title={Vlp: A survey on vision-language pre-training},
  author={Chen, Fei-Long and Zhang, Du-Zhen and Han, Ming-Lun and Chen, Xiu-Yi and Shi, Jing and Xu, Shuang and Xu, Bo},
  journal={Machine Intelligence Research},
  volume={20},
  number={1},
  pages={38--56},
  year={2023},
  publisher={Springer}
}

@inproceedings{wang2023image,
  title={Image as a foreign language: Beit pretraining for vision and vision-language tasks},
  author={Wang, Wenhui and Bao, Hangbo and Dong, Li and Bjorck, Johan and Peng, Zhiliang and Liu, Qiang and Aggarwal, Kriti and Mohammed, Owais Khan and Singhal, Saksham and Som, Subhojit and others},
  booktitle={Proceedings of the IEEE/CVF Conference on Computer Vision and Pattern Recognition},
  pages={19175--19186},
  year={2023}
}

@article{dou2022coarse,
  title={Coarse-to-fine vision-language pre-training with fusion in the backbone},
  author={Dou, Zi-Yi and Kamath, Aishwarya and Gan, Zhe and Zhang, Pengchuan and Wang, Jianfeng and Li, Linjie and Liu, Zicheng and Liu, Ce and LeCun, Yann and Peng, Nanyun and others},
  journal={Advances in neural information processing systems},
  volume={35},
  pages={32942--32956},
  year={2022}
}

@article{wan2025wan,
  title={Wan: Open and advanced large-scale video generative models},
  author={Wan, Team and Wang, Ang and Ai, Baole and Wen, Bin and Mao, Chaojie and Xie, Chen-Wei and Chen, Di and Yu, Feiwu and Zhao, Haiming and Yang, Jianxiao and others},
  journal={arXiv preprint arXiv:2503.20314},
  year={2025}
}

@article{kong2024hunyuanvideo,
  title={Hunyuanvideo: A systematic framework for large video generative models},
  author={Kong, Weijie and Tian, Qi and Zhang, Zijian and Min, Rox and Dai, Zuozhuo and Zhou, Jin and Xiong, Jiangfeng and Li, Xin and Wu, Bo and Zhang, Jianwei and others},
  journal={arXiv preprint arXiv:2412.03603},
  year={2024}
}

@article{tang2025human,
  title={Human-centric foundation models: Perception, generation and agentic modeling},
  author={Tang, Shixiang and Wang, Yizhou and Chen, Lu and Wang, Yuan and Peng, Sida and Xu, Dan and Ouyang, Wanli},
  journal={arXiv preprint arXiv:2502.08556},
  year={2025}
}

@article{liu2024sora,
  title={Sora: A review on background, technology, limitations, and opportunities of large vision models},
  author={Liu, Yixin and Zhang, Kai and Li, Yuan and Yan, Zhiling and Gao, Chujie and Chen, Ruoxi and Yuan, Zhengqing and Huang, Yue and Sun, Hanchi and Gao, Jianfeng and others},
  journal={arXiv preprint arXiv:2402.17177},
  year={2024}
}

@article{hacohen2024ltx,
  title={Ltx-video: Realtime video latent diffusion},
  author={HaCohen, Yoav and Chiprut, Nisan and Brazowski, Benny and Shalem, Daniel and Moshe, Dudu and Richardson, Eitan and Levin, Eran and Shiran, Guy and Zabari, Nir and Gordon, Ori and others},
  journal={arXiv preprint arXiv:2501.00103},
  year={2024}
}

@misc{genmo2024mochi,
      title={Mochi 1},
      author={Genmo Team},
      year={2024},
      publisher = {GitHub},
      journal = {GitHub repository},
      howpublished={\url{https://github.com/genmoai/models}}
}

@article{ma2024latte,
  title={Latte: Latent diffusion transformer for video generation},
  author={Ma, Xin and Wang, Yaohui and Jia, Gengyun and Chen, Xinyuan and Liu, Ziwei and Li, Yuan-Fang and Chen, Cunjian and Qiao, Yu},
  journal={arXiv preprint arXiv:2401.03048},
  year={2024}
}

@article{yang2024cogvideox,
  title={Cogvideox: Text-to-video diffusion models with an expert transformer},
  author={Yang, Zhuoyi and Teng, Jiayan and Zheng, Wendi and Ding, Ming and Huang, Shiyu and Xu, Jiazheng and Yang, Yuanming and Hong, Wenyi and Zhang, Xiaohan and Feng, Guanyu and others},
  journal={arXiv preprint arXiv:2408.06072},
  year={2024}
}

@article{wang2023modelscope,
  title={Modelscope text-to-video technical report},
  author={Wang, Jiuniu and Yuan, Hangjie and Chen, Dayou and Zhang, Yingya and Wang, Xiang and Zhang, Shiwei},
  journal={arXiv preprint arXiv:2308.06571},
  year={2023}
}

@article{zhang2025show,
  title={Show-1: Marrying pixel and latent diffusion models for text-to-video generation},
  author={Zhang, David Junhao and Wu, Jay Zhangjie and Liu, Jia-Wei and Zhao, Rui and Ran, Lingmin and Gu, Yuchao and Gao, Difei and Shou, Mike Zheng},
  journal={International Journal of Computer Vision},
  volume={133},
  number={4},
  pages={1879--1893},
  year={2025},
  publisher={Springer}
}

@article{wang2025lavie,
  title={Lavie: High-quality video generation with cascaded latent diffusion models},
  author={Wang, Yaohui and Chen, Xinyuan and Ma, Xin and Zhou, Shangchen and Huang, Ziqi and Wang, Yi and Yang, Ceyuan and He, Yinan and Yu, Jiashuo and Yang, Peiqing and others},
  journal={International Journal of Computer Vision},
  volume={133},
  number={5},
  pages={3059--3078},
  year={2025},
  publisher={Springer}
}

@article{guo2023animatediff,
  title={Animatediff: Animate your personalized text-to-image diffusion models without specific tuning},
  author={Guo, Yuwei and Yang, Ceyuan and Rao, Anyi and Liang, Zhengyang and Wang, Yaohui and Qiao, Yu and Agrawala, Maneesh and Lin, Dahua and Dai, Bo},
  journal={arXiv preprint arXiv:2307.04725},
  year={2023}
}

@article{unterthiner2019fvd,
  title={FVD: A new metric for video generation},
  author={Unterthiner, Thomas and Van Steenkiste, Sjoerd and Kurach, Karol and Marinier, Rapha{\"e}l and Michalski, Marcin and Gelly, Sylvain},
  year={2019}
}

@inproceedings{hessel2021clipscore,
  title={Clipscore: A reference-free evaluation metric for image captioning},
  author={Hessel, Jack and Holtzman, Ari and Forbes, Maxwell and Le Bras, Ronan and Choi, Yejin},
  booktitle={Proceedings of the 2021 conference on empirical methods in natural language processing},
  pages={7514--7528},
  year={2021}
}

@article{liu2023fetv,
  title={Fetv: A benchmark for fine-grained evaluation of open-domain text-to-video generation},
  author={Liu, Yuanxin and Li, Lei and Ren, Shuhuai and Gao, Rundong and Li, Shicheng and Chen, Sishuo and Sun, Xu and Hou, Lu},
  journal={Advances in Neural Information Processing Systems},
  volume={36},
  pages={62352--62387},
  year={2023}
}

@inproceedings{huang2024vbench,
  title={Vbench: Comprehensive benchmark suite for video generative models},
  author={Huang, Ziqi and He, Yinan and Yu, Jiashuo and Zhang, Fan and Si, Chenyang and Jiang, Yuming and Zhang, Yuanhan and Wu, Tianxing and Jin, Qingyang and Chanpaisit, Nattapol and others},
  booktitle={Proceedings of the IEEE/CVF Conference on Computer Vision and Pattern Recognition},
  pages={21807--21818},
  year={2024}
}

@inproceedings{ke2021musiq,
  title={Musiq: Multi-scale image quality transformer},
  author={Ke, Junjie and Wang, Qifei and Wang, Yilin and Milanfar, Peyman and Yang, Feng},
  booktitle={Proceedings of the IEEE/CVF international conference on computer vision},
  pages={5148--5157},
  year={2021}
}

@inproceedings{caron2021emerging,
  title={Emerging properties in self-supervised vision transformers},
  author={Caron, Mathilde and Touvron, Hugo and Misra, Ishan and J{\'e}gou, Herv{\'e} and Mairal, Julien and Bojanowski, Piotr and Joulin, Armand},
  booktitle={Proceedings of the IEEE/CVF international conference on computer vision},
  pages={9650--9660},
  year={2021}
}

@article{zheng2025vbench,
  title={Vbench-2.0: Advancing video generation benchmark suite for intrinsic faithfulness},
  author={Zheng, Dian and Huang, Ziqi and Liu, Hongbo and Zou, Kai and He, Yinan and Zhang, Fan and Gu, Lulu and Zhang, Yuanhan and He, Jingwen and Zheng, Wei-Shi and others},
  journal={arXiv preprint arXiv:2503.21755},
  year={2025}
}

@article{yuan2024chronomagic,
  title={Chronomagic-bench: A benchmark for metamorphic evaluation of text-to-time-lapse video generation},
  author={Yuan, Shenghai and Huang, Jinfa and Xu, Yongqi and Liu, Yaoyang and Zhang, Shaofeng and Shi, Yujun and Zhu, Rui-Jie and Cheng, Xinhua and Luo, Jiebo and Yuan, Li},
  journal={Advances in Neural Information Processing Systems},
  volume={37},
  pages={21236--21270},
  year={2024}
}

@inproceedings{sun2025t2v,
  title={T2v-compbench: A comprehensive benchmark for compositional text-to-video generation},
  author={Sun, Kaiyue and Huang, Kaiyi and Liu, Xian and Wu, Yue and Xu, Zihan and Li, Zhenguo and Liu, Xihui},
  booktitle={Proceedings of the Computer Vision and Pattern Recognition Conference},
  pages={8406--8416},
  year={2025}
}

@inproceedings{wang2025your,
  title={Is your world simulator a good story presenter? a consecutive events-based benchmark for future long video generation},
  author={Wang, Yiping and He, Xuehai and Wang, Kuan and Ma, Luyao and Yang, Jianwei and Wang, Shuohang and Du, Simon Shaolei and Shen, Yelong},
  booktitle={Proceedings of the Computer Vision and Pattern Recognition Conference},
  pages={13629--13638},
  year={2025}
}

@article{qiao2025vadb,
  title={VADB: A Large-Scale Video Aesthetic Database with Professional and Multi-Dimensional Annotations},
  author={Qiao, Qianqian and Zheng, DanDan and Bo, Yihang and Peng, Bao and Huang, Heng and Jiang, Longteng and Wang, Huaye and Chen, Jingdong and Zhou, Jun and Jin, Xin},
  journal={arXiv preprint arXiv:2510.25238},
  year={2025}
}

@article{matbouly2022quantifying,
  title={Quantifying the unquantifiable: the color of cinematic lighting and its effect on audience’s impressions towards the appearance of film characters},
  author={Matbouly, Mustafa Yousry},
  journal={Current Psychology},
  volume={41},
  number={6},
  pages={3694--3715},
  year={2022},
  publisher={Springer}
}

@article{deren1960cinematography,
  title={Cinematography: the creative use of reality},
  author={Deren, Maya},
  journal={Daedalus},
  volume={89},
  number={1},
  pages={150--167},
  year={1960},
  publisher={JSTOR}
}

@book{brown2016cinematography,
  title={Cinematography: theory and practice: image making for cinematographers and directors},
  author={Brown, Blain},
  year={2016},
  publisher={Routledge}
}

@article{Qwen3-VL,
      title={Qwen3-VL Technical Report}, 
      author={Shuai Bai and Yuxuan Cai and Ruizhe Chen and Keqin Chen and Xionghui Chen and Zesen Cheng and Lianghao Deng and Wei Ding and Chang Gao and Chunjiang Ge and Wenbin Ge and Zhifang Guo and Qidong Huang and Jie Huang and Fei Huang and Binyuan Hui and Shutong Jiang and Zhaohai Li and Mingsheng Li and Mei Li and Kaixin Li and Zicheng Lin and Junyang Lin and Xuejing Liu and Jiawei Liu and Chenglong Liu and Yang Liu and Dayiheng Liu and Shixuan Liu and Dunjie Lu and Ruilin Luo and Chenxu Lv and Rui Men and Lingchen Meng and Xuancheng Ren and Xingzhang Ren and Sibo Song and Yuchong Sun and Jun Tang and Jianhong Tu and Jianqiang Wan and Peng Wang and Pengfei Wang and Qiuyue Wang and Yuxuan Wang and Tianbao Xie and Yiheng Xu and Haiyang Xu and Jin Xu and Zhibo Yang and Mingkun Yang and Jianxin Yang and An Yang and Bowen Yu and Fei Zhang and Hang Zhang and Xi Zhang and Bo Zheng and Humen Zhong and Jingren Zhou and Fan Zhou and Jing Zhou and Yuanzhi Zhu and Ke Zhu},
	  journal={arXiv preprint arXiv:2511.21631},
      year={2025}
}

@article{liu2025flow,
  title={Flow-grpo: Training flow matching models via online rl},
  author={Liu, Jie and Liu, Gongye and Liang, Jiajun and Li, Yangguang and Liu, Jiaheng and Wang, Xintao and Wan, Pengfei and Zhang, Di and Ouyang, Wanli},
  journal={arXiv preprint arXiv:2505.05470},
  year={2025}
}

@article{jiang2026vga,
  title={VGA-Bench: A Unified Benchmark and Multi-Model Framework for Video Aesthetics and Generation Quality Evaluation},
  author={Jiang, Longteng and Zheng, DanDan and Qiao, Qianqian and Huang, Heng and Wang, Huaye and Bo, Yihang and Peng, Bao and Chen, Jingdong and Zhou, Jun and Jin, Xin},
  journal={arXiv preprint arXiv:2604.10127},
  year={2026}
}

\end{document}